%% file: main.tex
\documentclass[sigconf]{acmart}

\AtBeginDocument{%
  }

\copyrightyear{2025}
\acmYear{2025}
\setcopyright{rightsretained}
\acmConference[KDD '25] {Proceedings of the 1st Workshop on "AI for Supply Chain: Today and Future" @ 31st ACM SIGKDD Conference on Knowledge Discovery and Data Mining V.2}{August 3, 2025}{Toronto, ON, Canada.}
\acmBooktitle{Proceedings of the 1st Workshop on "AI for Supply Chain: Today and Future" @ 31st ACM SIGKDD Conference on Knowledge Discovery and Data Mining V.2 (KDD '25), August 3, 2025, Toronto, ON, Canada}
\acmISBN{979-8-4007-1454-2/25/08}
\acmDOI{10.1145/XXXXXX.XXXXXX}
\usepackage{amsmath}
\usepackage{bm}
\usepackage{xspace}
\usepackage{multirow}
\usepackage{hyperref}
\usepackage{subcaption}
\usepackage{pifont}
\usepackage{bbm}
\usepackage{makecell}
\usepackage{balance}

\newcommand{\method}{TAT\xspace}
\newcommand{\methodns}{TAT}
\newcommand{\att}{Temporal Alignment Attention\xspace}
\newcommand{\taa}{TAA}

\begin{document}

\title{TAT: Temporal-Aligned Transformer for Multi-Horizon \\ Peak Demand Forecasting}

\author{Zhiyuan Zhao$^*$}
\email{leozhao1997@gatech.edu}
\affiliation{%
\institution{Georgia Institute of Technology}
\city{Atlanta}
\state{GA}
\country{USA}
}

\author{Sitan Yang$^*$}
\email{syang@keystonestrategy.com}
\affiliation{%
\institution{Keystone AI}
\city{New York}
\state{NY}
\country{USA}
}

\author{Kin G. Olivares}
\email{kigutie@amazon.com}
\affiliation{%
 \institution{Amazon}
 \city{New York}
 \state{NY}
 \country{USA}
}

\author{Boris N. Oreshkin}
\email{oreshkin@amazon.com}
\affiliation{%
\institution{Amazon}
\city{New York}
\state{NY}
\country{USA}
}

\author{Stan Vitebsky}
\email{vitebsky@amazon.com}
\affiliation{%
 \institution{Amazon}
 \city{New York}
 \state{NY}
 \country{USA}
}

\author{Michael W. Mahoney}
\email{zmahmich@amazon.com}
\affiliation{%
\institution{Amazon}
\city{New York}
\state{NY}
\country{USA}
}

\author{B. Aditya Prakash}
\email{badityap@cc.gatech.edu}
\affiliation{%
\institution{Georgia Institute of Technology}
\city{Atlanta}
\state{GA}
\country{USA}
}

\author{Dmitry Efimov}
\email{defimov@amazon.com}
\affiliation{%
\institution{Amazon}
\city{New York}
\state{NY}
\country{USA}
}


\renewcommand{\shortauthors}{Z. Zhao et al.}


\begin{abstract}
Multi-horizon time series forecasting has many practical applications such as demand forecasting. Accurate demand prediction is critical to help make buying and inventory decisions for supply chain management of e-commerce and physical retailers, and such predictions are typically required for future horizons extending tens of weeks.
This is especially challenging during high-stake sales events when demand peaks are particularly difficult to predict accurately. 
However, these events are important not only for managing supply chain operations but also for ensuring a seamless shopping experience for customers. 
To address this challenge, we propose Temporal-Aligned Transformer (\method), a multi-horizon forecaster leveraging apriori-known context variables such as holiday and promotion events information for improving predictive performance. 
Our model consists of an encoder and decoder, both embedded with a novel \att (\taa), designed to learn context-dependent alignment for peak demand forecasting. 
We conduct extensive empirical analysis on two large-scale proprietary datasets from a large e-commerce retailer.
We demonstrate that \method brings up to 30\% accuracy improvement on peak demand forecasting while maintaining competitive overall performance compared to other state-of-the-art methods.

\renewcommand{\thefootnote}{\fnsymbol{footnote}}
\footnotetext[1]{This work was done when affiliated with Amazon, USA.}
\renewcommand{\thefootnote}{\arabic{footnote}}  

\end{abstract}





\maketitle

\input{content/01_intro}
\input{content/03_problem}
\input{content/04_method}

\input{content/05_exp}

\input{content/06_conclusion}

\bibliographystyle{ACM-Reference-Format}
\bibliography{reference}

\appendix
\input{content/02_relate}
\input{content/08_appb}
\input{content/09_appc}

\end{document}

%% file: content/01_intro.tex
\section{Introduction}
\label{sec:intro}


Time series forecasting is a fundamental problem that finds broad applications in supply chain management, finance, healthcare, and more~\cite{granger2014forecasting, babai2022demand, mathis2024title}. 
In modern forecasting problems, predictions are often required over multiple time horizons, a task known as multi-horizon forecasting~\cite{wen2017multi,eisenach2020mqtransformer}. 
Recent advances in deep learning time series models have achieved notable success, exemplified by recurrent neural networks and transformer-based architectures~\cite{lai2018modeling, oreshkin2019n, zhou2021informer, Yuqietal-2023-PatchTST, zhao2025timerecipe}. 
These methods have already demonstrated state-of-the-art predictive performance for producing accurate forecasts of horizons up to tens of weeks in the future. 

As a practical motivation for this work, we consider industrial-scale demand prediction at a large e-commerce retailer, where forecasts are essential to inform critical supply chain decisions~\cite{chen2000quantifying, bose2017probabilistic}. 
One key challenge is to accurately predict demand patterns during peak sales events, a problem becoming increasingly important as the number and scale of such events have grown dramatically in recent years. 
Marked by an outstanding concentration of merchandising discounts and promotions, these peak events differ distinctively from typical factors like seasonality or product life cycles, resulting in hard-to-predict demand spikes. 
Both over-forecasting and under-forecasting peak demand can be problematic, leading to poor customer experience and large financial costs. 

General time series methods that only focus on univariate forecasting~\cite{oreshkin2019n} or that treat multivariate forecasting as a channel independent setup~\cite{Yuqietal-2023-PatchTST} are very unlikely to correctly account for the critical impact of key exogenous variables, such as promotion and price-related information, making them ineffective for practical demand forecasting tasks. 
To address these limitations, the MQ-Forecaster framework~\cite{wen2017multi} was introduced with designs suited for demand forecasting. It incorporates common exogenous variables such as seasonal cycles and planned peak events, but its architecture suffers from an ``information bottleneck,'' where the encoder transmits information to the decoder only via a single hidden state. 
There are many subsequent works leveraging the celebrated Transformer architecture~\cite{vaswani2017attention} that aim to address this issue. 
One such example is Temporal Fusion Transformer (TFT)~\cite{lim2021temporal}, which has been shown to achieve competitive performance for a few commonly used public datasets. 
However, despite being general, TFT simply concatenates all features and applies self-attention. 
This attention mechanism can work reasonably well for small yet simple datasets.
As we show later in this paper, however, it fails to perform similarly well in real demand forecasting tasks, mainly due to the failure to correctly leverage the power of key exogenous variables. 

Given the limitations and challenges, in this paper, 
we introduce Temporal-Aligned Transformer (\method), a novel framework that is explicitly formulated to enhance learning between the target time series and key exogenous variables, providing the context on which the model produces the forecast. 
Since the main application of this paper is peak demand forecasting, we consider exogenous variables related to future peak events. 
Our approach proves to enhance peak demand prediction accuracy, while maintaining competitive forecasting performance during non-peak periods, compared to popular forecasting methods. 
The main contributions are summarized as follows:
\begin{itemize}
    \item 
    We introduce \att (TAA), a novel method for learning the complex dependencies between the target time series and key exogenous variables which provide the context information, and thus enable more accurate forecasts as compared to the widely adopted self-attention mechanisms in previous methods. 
    \item 
    We propose Temporal-Aligned Transformer (\method), a Transformer based model that integrates TAA within an encoder-decoder architecture and demonstrates the SOTA performance in multi-horizon time series forecasting.  
    
    \item 
    We perform comprehensive evaluations of TAT against popular forecasting methods
    on two substantially complex industrial demand datasets from a large e-commerce retailer, we achieve up to 30\% improvements over the previously reported SOTA in peak demand prediction, while still maintaining competitive results in overall performance.
\end{itemize}
We provide more detailed related works in Appendix~\ref{sec:relate}.

%% file: content/03_problem.tex
\section{Problem Formulation}
\label{sec:problem}

We adopt the same setting as in the TFT method~\cite{lim2021temporal}, and we formulate the demand forecasting problem as an application of multi-horizon time-series forecasting, where the objective is to predict the target time series by leveraging multiple types of features. 
Formally, we define the dataset $\mathcal{D}$ consisting of $N$ samples over time steps up to $T$. In general,
each sample includes three types of features: 
static (time-invariant) metadata information $x^s \in \mathbb{R}^{d_s}$ (e.g., product category); 
historical observed time series covariates $\bm{x}^b \in \mathbb{R}^{T \times d_b}$ (e.g, product detail page views); and 
time-varying known context about the future $\bm{x}^c \in \mathbb{R}^{(T+H) \times d_c}$ (e.g., indicators for historical and upcoming holiday dates). 
In line with TFT and other direct methods~\cite{wen2017multi, eisenach2020mqtransformer}, we simultaneously predict the future values of $y \in \mathbb{R}^H$ for $H$ time steps, i.e., $h\in\{1,...,H\}$. We incorporate static information $x_s$ and all historical observed information $\bm{x}^b$ with a lookback window $L$ (i.e., $\bm{x}^b_{t-L:t} = [x^b_{t-L}, \ldots, x^b_L]$). 
For known context time series $\bm{x}^c$, we additionally leverage its entire range of time steps (i.e., $\bm{x}^c_{t-L:t+H} = [x^c_{t-L}, \ldots, x^c_{t+H}]$).  
For simplicity, we denote the time range from $t-L$ to $t$ with a down script $[L]$, the time range from $t+1$ to $t+H$ with $[H]$, and the whole lookback and horizon range as $[L+H]$. 
We parametrize the model by $\theta$, and we learn through empirical risk minimization, obtaining a function $f_{\theta}:(x^s, \bm{x}^b_{[L]}, \bm{x}^c_{[L+H]}) \rightarrow y_{[H]}$. 




\begin{figure}[t]
    \centering
    \includegraphics[width=0.8\linewidth]{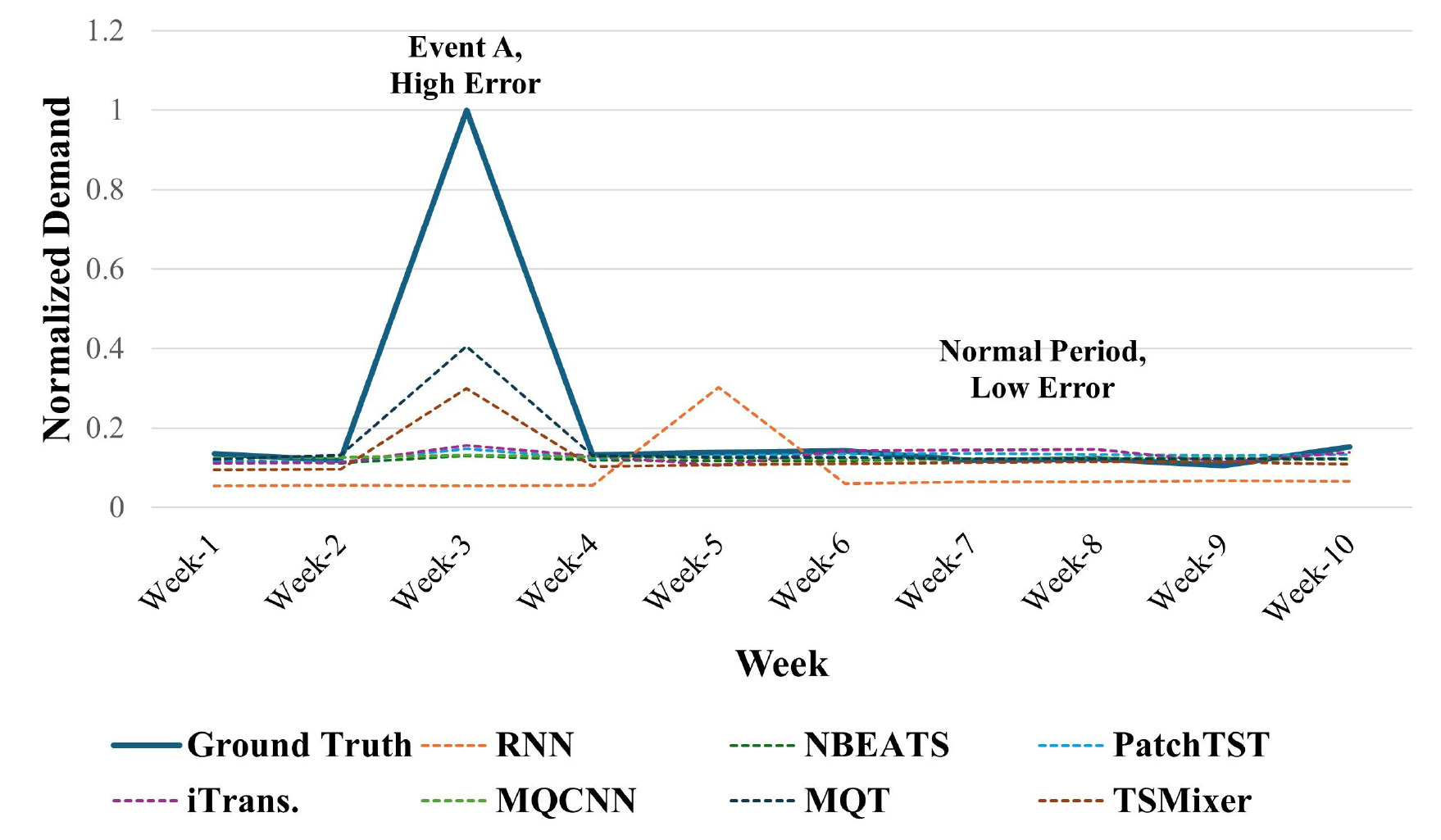}
    \caption{Week-by-week (normalized) demand for a popular book. Even with state-of-the-art forecasting models, the forecasting errors remain high during demand peaks.
    }
    \label{fig:asin}
\end{figure}

Existing forecasting approaches~\cite{wen2017multi, Olivares2021ProbabilisticHF, eisenach2020mqtransformer, lim2021temporal, wolff2024spade} have achieved some success in overall forecasting accuracy. 
However, these approaches often struggle when predicting demand peaks, particularly during holidays and promotional events, as showcased in Figure~\ref{fig:asin}. 
In this paper, we emphasize the importance of accurately predicting demand peaks, and we highlight the critical impact of known context features, \( \bm{x}^c \), in improving model performance for capturing demand patterns during peak events. 
These exogenous variables are broadly available, and they have been considered in previous studies \cite{lim2021temporal,chen2023tsmixer}. 
They are often deterministic for both historical lookback and future horizon periods, and in the industrial demand forecasting problem, $\bm{x}^c$ typically contains additional key features related to promotion events such as price discount amount and promotion type. 
Our main motivation is to improve peak demand forecasting accuracy by effectively leveraging $x^c$.

%% file: content/04_method.tex
\section{Temporal-Aligned Transformer}
\label{sec:method}

In this section, we introduce the architecture of the proposed Temporal-Aligned Transformer (TAT) model. 
In particular,
We first describe the \att (TAA), which provides an explicit context-dependent learning for target demand time series based on the known context features related to demand peaks. 
We then introduce the architecture of \method that effectively integrates \att in its encoder-decoder structure.


\subsection{\att}
\label{sec:att}

We introduce \att (TAA), 
a module designed to temporally align demand patterns with known contextual variables related to demand peaks, while also capturing their interactions within both the encoder and decoder through attention mechanisms. 
Specifically, we consider $x^c$, the known contextual information comprising various holiday-related and promotion-related features, and we partition its time steps into two components: the historical part, $x^c_{[L]} \in \mathbb{R}^{L \times d_c}$, which is set to align with observed demand patterns $x^b_{[L]}$, and the apriori-known future horizon context, $x^c_{[H]} \in \mathbb{R}^{H \times d_c}$, which aligns with the forecasting horizon. 
This explicit separation is rarely addressed in existing demand forecasting methods.  

To achieve this temporal alignment, we leverage attention mechanisms in both the encoder and decoder with inputs from $e$, the latent embedding space. We denote the output of any input variable through the latent space with $e$, e.g., $\text{embed}(x^b_{[L]}) = e^b_{[L]}$. This alignment is achieved using the standard multi-head scaled dot-product attention~\cite{vaswani2017attention}, i.e.,  
$\text{TAA}(Q, K, V) = \text{MultiHead}(Q, K, V)$. Nevertheless,
we propose using selective features for attention inputs, rather than adopting the identical Q, K, and V setup of vanilla self-attention, as commonly used in previous forecasting models. 
In the encoder, we choose the embedded observed time series $e^b_{[L]}$ as Q to align with the embedded historical information of context variables $e^c_{[L]}$. 
For finer-grained distinctions, we include in K the embedded static features $e^s$ as metadata to differentiate demand characteristics across products. 
For example, some products such as routine utilities can be less sensitive to promotional events compared to items like electronic devices, a distinction that can be explicitly made by incorporating the metadata.
Lastly, we include all features in V to enable additional cross-series interactions to the attention output. Consequently, the Q, K, and V setup of TAA in the encoder is defined as follows:  
\begin{equation}
\begin{aligned}
    &Q=\text{Linear}(e^b_{[L]}), \\
    &K=\text{Linear}([e^c_{[L]}, \text{Broadcast(}e^s)]), \\
    &V=\text{Linear}([e^b_{[L]}, e^c_{[L]}, \text{Broadcast(}e^s)]),\\
    &\text{Score}(Q,K) \in \mathbb{R}^{L\times L} .
\end{aligned}
\end{equation}
Here, \([\cdot]\) denotes concatenation along the hidden dimension, resulting in different dimensionalities for Q, K, and V. We then project all tensors onto a unified hidden dimension $d_{hidden}$ using linear layers for attention computation. Additionally, the static features $e^s$ are broadcasted to match the temporal dimension of the other input sequences.  

Similarly, in the decoder, we use the initialization sequence $e^\text{Dec}_{[H]}$, derived from the encoder output, as Q to represent demand forecasts across all future horizons. These forecasts are then aligned with the embedded known future context variables $e^c_{[H]}$ through an additional TAA mechanism, with the Q, K, and V setup defined as:  
\begin{equation}
\begin{aligned}
    &Q=\text{Linear}(e^\text{Dec}_{[H]}), \\
    &K=\text{Linear}([e^c_{[H]}, \text{Broadcast(}e^s)]), \\
    &V=\text{Linear}([e^b_{[H]}, e^c_{[H]}, \text{Broadcast(}e^s)]),\\
    &\text{Score}(Q,K) \in \mathbb{R}^{H\times H} .
\end{aligned}
\end{equation}

The primary objective is to enhance context alignment within the attention mechanism by designing Q, K, and V. 
This enables the network to learn temporal alignment in both lookback and horizon windows by assigning greater weights to specific positions, namely those where $x^b$ exhibits demand peaks and contextual features indicate relevant signals.
In contrast, TFT~\cite{lim2021temporal} uses standard multi-head self-attention across all types of feature embeddings.
This can be too general to capture the intricate interactions during peak events. 
MQTransformer~\cite{eisenach2020mqtransformer} has a similar attention module for context features $x^c$. 
However, it compresses $x^c$ along the temporal dimension and only applies attention between the observed target time series and the future known horizon context, causing misalignment between these features.

\begin{figure*}[t]
    \centering
    \includegraphics[width=0.8\linewidth]{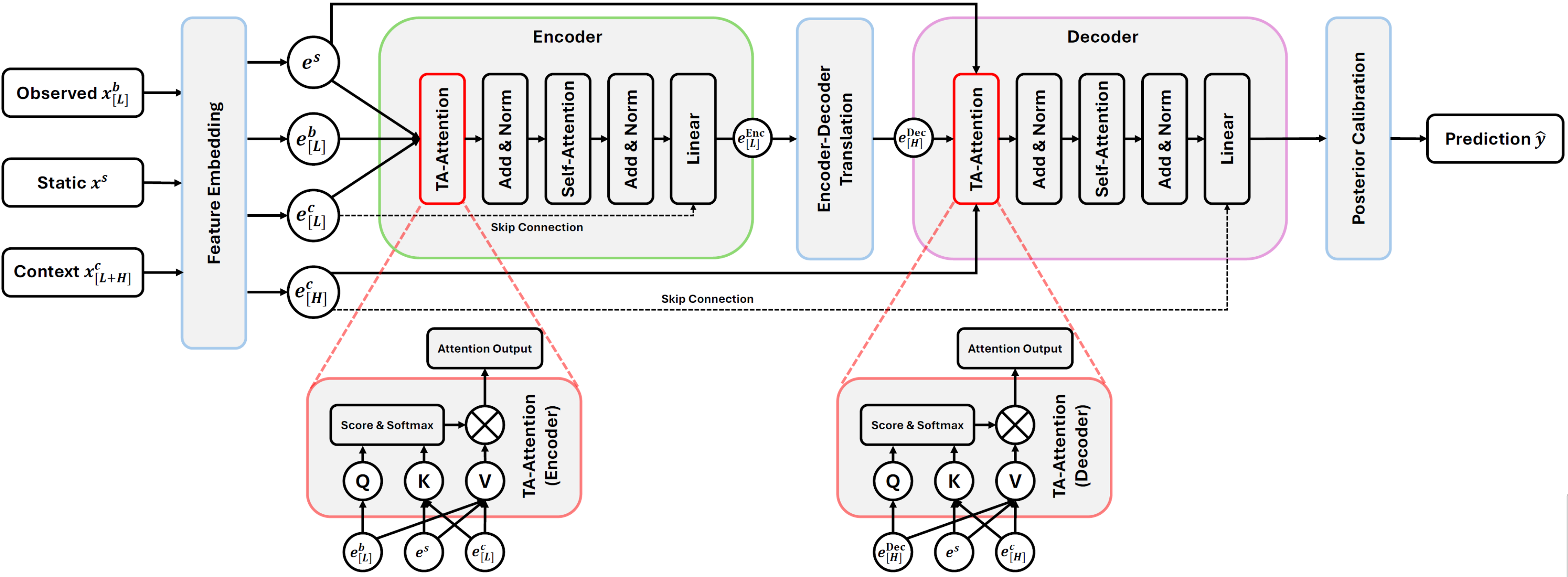}
    \Description{}
    \caption{\method model architecture, with the proposed TAA highlighted in red. 
    }
    \label{fig:tat}
\end{figure*}

\subsection{TAT Model Architecture}
\label{sec:model}

The model architecture of TAT is shown in Figure~\ref{fig:tat} and consists of the following components: the input feature embedding module; the encoder module; the encoder-decoder translation module; the decoder module; and the posterior calibration module. 
Each module performs distinct functions sequentially. 
The feature embedding module (Section~\ref{sxn:tat-feature-embedding}) applies different yet effective embeddings for various feature types.
This is followed by the encoder (Section~\ref{sxn:tat-enc-dec}), which incorporates a TAA together with self-attention to effectively align and enhance the learning between the target time series and promotion-related features. 
Next, an encoder-decoder translation (Section~\ref{sxn:tat-trans}) with self-attention generates the decoder sequence based on historical patterns.
Then, the decoder applies TAA and self-attention similar to those in the encoder. 
Finally, posterior calibration (Section~\ref{sxn:tat-calib}) further refines the predictions during demand peaks. 
The overall model architecture is illustrated in Figure~\ref{fig:tat}, and the details of each component are provided in the following subsections.


\subsubsection{\textbf{Input Feature Embedding.}} 
\label{sxn:tat-feature-embedding}

\begin{figure}[h]
    \centering
    \includegraphics[width=0.9\linewidth]{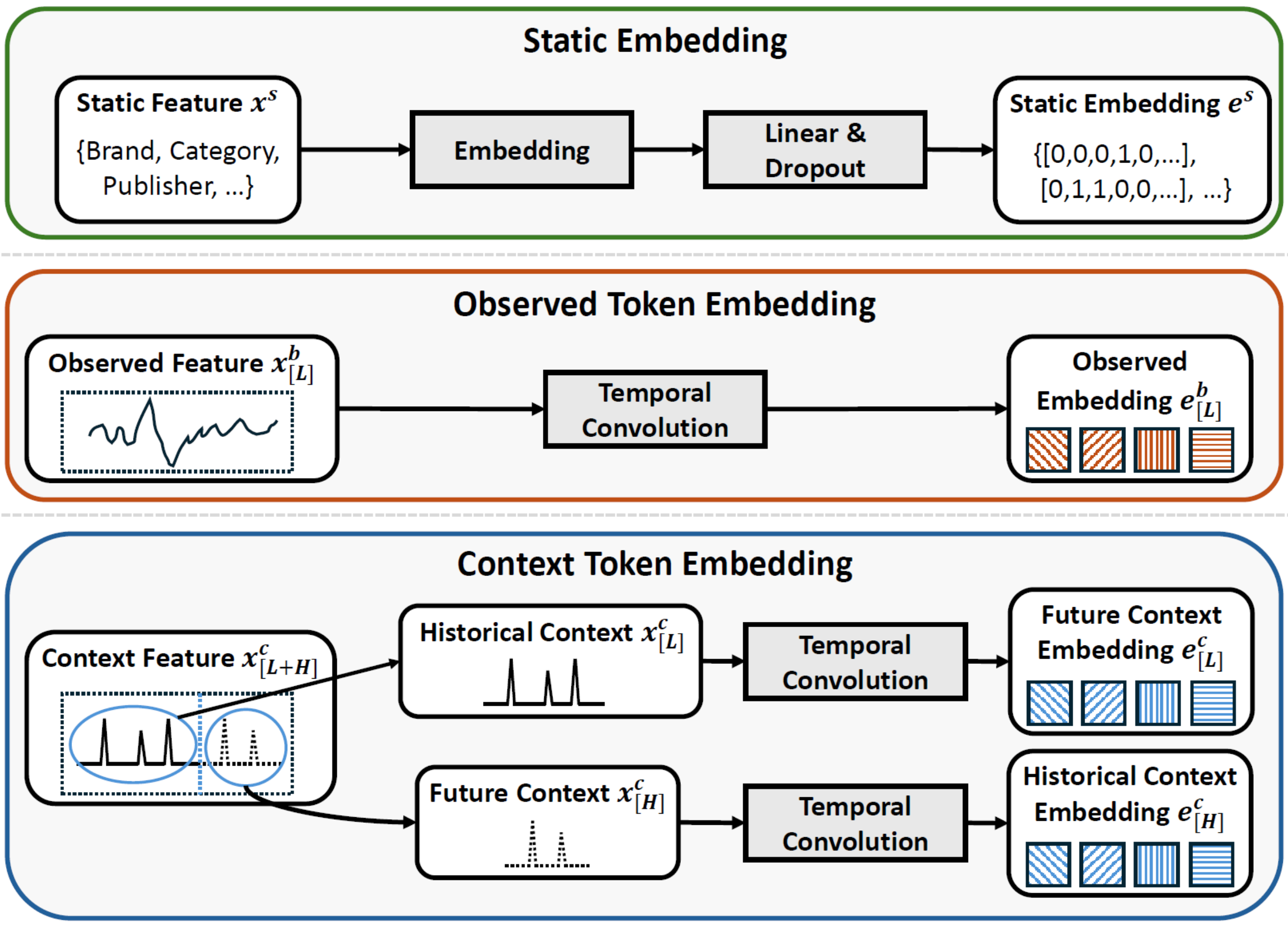}
    \caption{Diagram of the input feature embedding process for static, observed, and known context features. 
    }
    \label{fig:emb}
\end{figure}

We apply different strategies to transform each type of input feature into embedding vectors, as shown in Figure~\ref{fig:emb}.
For static features \(x^s \in \mathbb{R}^s\), we use categorical embedding to map them into a high-dimensional latent space, followed by dropout and a linear projection to a fixed hidden size. Subsequently, we broadcast them along the temporal dimension to be compatible with temporal features. 
For the observed historical time series \(x^b_{[L]} \in \mathbb{R}^{L \times d_b}\), we apply token embedding via standard 1D dilated convolution layers \cite{wavenet16} across time steps, followed by embedding strategies for time series, as described in previous work~\cite{zhou2021informer}. 
For known context features \(x^c_{[L+H]} \in \mathbb{R}^{(L+H)\times d_c}\), we split these time series into historical (\(x^c_{[L]} \in \mathbb{R}^{L \times d_c}\)) and future (\(x_{[H]}^c\in\mathbb{R}^{H \times d_c}\)) components. We then perform separate token embedding as on each part, similar to the approach used for \(x^b\). 

\subsubsection{\textbf{Encoder-Decoder Structure.}} 
\label{sxn:tat-enc-dec}
\method follows an encoder-decoder architecture, as does a vanilla Transformer~\cite{vaswani2017attention}, but it employs a customized attention module in both the encoder and decoder. 
This module consists of two attention layers: TAA; followed by a regular self-attention. 
TAA first aligns the exogenous context promotion-related features with the target time series (as described in Section~\ref{sec:att}); and the self-attention uses the identical query, key, and value derived from the TAA outputs, to refine the aligned representations, thereby improving the overall representation quality. To ensure stable training for the attention mechanisms, we apply layer normalization and residual connections after the attention outputs, same as the vanilla Transformer~\cite{vaswani2017attention} and Informer~\cite{zhou2021informer}. 
The entire processing module described above forms the core structure of both the encoder and decoder of \method.

\subsubsection{\textbf{Encoder-Decoder Translation.}} 
\label{sxn:tat-trans}
To generate the decoder initialization sequence (i.e., the embedding of the decoder's input), $e_{[H]}^\text{Dec}$, from the encoder output tokens, we employ an encoder-decoder translation module to map the encoder output to the decoder input, achieved by a transposed self-attention. 
This self-attention uses the transposed encoder tokens as input, with the temporal dimension placed as the final dimension. The self-attention approach treats each hidden dimension as a token and predicts the future tokens through modeling channel-wise dependencies, following a methodology similar to iTransformer~\cite{liu2023itransformer}.
It then has a linear projection layer to adjust the token size from the lookback window size to the horizon window size, generating the initial latent representation for the decoder. 
The full operation is as follows:
\begin{equation}
\begin{aligned}
    e_{[L]}^{\text{Enc}} &= \text{Encoder}(e_{[L]}^b, e^c_{[L]}, e^s) \in \mathbb{R}^{L\times d_{hidden}}\\
    e_{[H]}^{\text{Dec}} &= 
    \text{Linear}(\text{Self-Attention}(Q,K,V))^T \in \mathbb{R}^{H\times d_{hidden}} \\
    &\text{where} \: Q,K, V=(e_{[L]}^{\text{Enc}})^T ,
\end{aligned}
\end{equation}
where \(e_{[L]}^{\text{Enc}}\) denotes the encoded tokens from the encoder output.  

Previous studies~\cite{liu2023itransformer} typically treat $e_{[H]}^{\text{Dec}}$ as the final predictions, but we propose leveraging these predictions as the decoder initialization. This approach further enhances the alignment between predicted patterns and the known future context variables $x^c_{[H]}$, allowing refinement through temporal alignment in the decoder.  
Moreover, compared to the commonly used zero decoder initialization in time-series forecasting~\cite{zhou2021informer}, the encoder-decoder self-attention enables more effective use of historical patterns, and also avoid the information bottleneck described in \cite{eisenach2020mqtransformer} due to transmitting from encoder to decoder via a single hidden state. 
By reshaping and projecting past information onto future horizons, this mechanism generates a more effective decoder initialization sequence.  

\subsubsection{\textbf{Posterior Calibration.}} 
\label{sxn:tat-calib}
Besides the model accuracy, calibration issues often arise during demand peaks, as it is generally hard to predict the magnitude of impact on demand with a concentration of drivers. Therefore,
in TAT, we design an additional step referred to as the Posterior Calibration module, as follows:
\begin{equation}
\begin{gathered}
    y_{[H]}^{\text{Dec}} = \text{Decoder}(e^{\text{Dec}}_{[L]}, e^c_{[H]}, e^s)\\
    \hat{y} = y_{[H]}^{\text{Dec}} * (1 + \text{Calib}(x^c_{[H]})) ,
\end{gathered}
\end{equation}
where $y^{\text{Dec}}_{[H]}$ is the decoder's output, $\hat{y}$ is the final calibrated prediction, and $\text{Calib}(\cdot)$ is a simple multi-layer perceptron (MLP) serving as the calibration layer. This layer takes future features $x^c$ as input and outputs a scaling factor to adjust the decoder's predictions. The rationale behind the calibration is that predictions during demand peaks are generally more challenging compared to normal periods.
Posterior calibration aims to rescale predictions at demand peaks, while preserving the accuracy of predictions for normal periods. 

\subsubsection{\textbf{Loss Function.}} 
\method is trained by jointly minimizing a set of quantile losses. 
Consistent with previous studies, we evaluate the $50^\text{th}$ and $90^\text{th}$ quantile losses (unweighted $P_{50}$ and $P_{90}$), commonly used in forecasting to provide predictions with associated uncertainty and is employed in popular literature~\cite{wen2017multi,yu2016temporal,eisenach2020mqtransformer}. 
The overall loss function is formulated as follows:
\begin{equation}
\begin{aligned}
    \mathcal{L} &= \mathcal{L}_{P50}(y, \hat{y}^1) + \mathcal{L}_{P90}(y, \hat{y}^2) \\
    &= \sum \Big[ 0.5 \max(y - \hat{y}^1, 0) + 0.5 \max(\hat{y}^1 - y, 0) \Big] \\
    & + \sum \Big[ 0.9 \max(y - \hat{y}^2, 0) + 0.1 \max(\hat{y}^2 - y, 0) \Big]
    \label{eqn:ql}
\end{aligned}
\end{equation}
where \(y\) represents the ground truth values, \(\hat{y}^1\) denotes the $50^\text{th}$ quantile predictions, and \(\hat{y}^2\) corresponds to the $90^\text{th}$ quantile predictions. 
\method produces two-dimensional outputs from its final linear layer: the first dimension provides the $50^\text{th}$ quantile predictions (\(\hat{y}^1\)); and the second dimension provides the $90^\text{th}$ quantile predictions (\(\hat{y}^2\)). 
The training objective jointly optimizes the loss by summing both quantile-specific losses.

%% file: content/05_exp.tex
\input{tables/01_promotional_events.tex}

\section{Empirical results}
\label{sec:exp}


\subsection{Setup}

\subsubsection{Dataset.} 
We evaluate \method by comparing its performance against popular time-series forecasting baselines on two large-scale retail demand datasets from a large e-commerce retailer, namely \textbf{E-Commerce Retail Datasets}.
We obtain a large-scale proprietary demand data consisting of worldwide products sold on a large e-commerce retailer with demand time series for hundreds of thousands of products spanning more than 10 years.
These time series are in weekly grain and compounded with various exogenous features. 
With the data, we further select two smaller datasets, namely \textbf{E-Commerce Retail-Region 1} and \textbf{Region 2}, with products from two main geographic locations, where demand patterns can be quite different even for the same product. 
We use the 4-year period data for training and the following last year for testing. More detailed dataset statistics are in Table~\ref{tbl:stat}, Appendix~\ref{sec:appb}.

\subsubsection{Baselines.} 
For general time-series forecasting baselines, we include conventional RNN~\cite{chung2014empirical} and N-BEATS~\cite{oreshkin2019n, olivares2022nbeatsx}, and most recent SOTA forecasting models PatchTST~\cite{Yuqietal-2023-PatchTST}, iTransformer~\cite{liu2023itransformer} (iTrans.), and TSMixer~\cite{chen2023tsmixer}. 
(Univariate, multivariate, and univariate with exogenous features models, respectively.) 
For industrial demand forecasting baselines, we include MQCNN~\cite{wen2017multi}, TFT~\cite{lim2021temporal}, and MQTransformer (MQT)~\cite{eisenach2020mqtransformer}. 
Other well-known baselines, such as ARIMA~\cite{hyndman2024forecasting}, DeepAR~\cite{salinas2020deepar}, and ConvTrans~\cite{li2019enhancing}, are omitted since their performances are less competitive than the selected baselines shown by previous studies~\cite{lim2021temporal, eisenach2020mqtransformer}. 

\subsubsection{Evaluation Metrics.} 
We evaluate the performance of the baselines and our proposed \method using two metrics as described below.
The first metric is the standard \textit{overall accuracy} (Table~\ref{tbl:retail_all}), where we use the weighted $50^\text{th}$ and $90^\text{th}$ percentile quantile loss ($P_{50}$ and $P_{90}$)
~\cite{eisenach2020mqtransformer}, defined as follows:\
\begin{equation}
    P_{\alpha} = 2\frac{ \sum_{y_{i, t}\in \tilde{\Omega}} \sum^{H}_{h=1}\mathcal{L}_{\alpha}(y_{i, t}, \hat{y}(\alpha, h)) }{\sum_{y_{i, t}\in \tilde{\Omega}} \sum^{H}_{h=1} |y_{i, t}| }, \quad \alpha = 50,90 .
\label{eqn:metric}
\end{equation}
where $\Omega$ is the domain of training data containing all $N$ samples and $\mathcal{L}_{\alpha}$ is the quantile loss function described in Equation (\ref{eqn:ql}). 

In addition to overall accuracy,
We also consider \textit{target date accuracy} (Table~\ref{tbl:fixdate}) as an important metric as it measures the model performance directly towards the peak events of interest. In this case, $P_{50}$ and $P_{90}$ are defined as follows:
\begin{equation}
    P_{\alpha} = 2\frac{ \sum_{y_{i,t}\in \tilde{\Omega}} \sum^{H}_{h=1}\mathcal{L}_{\alpha}(y_{i,t}, \hat{y}(\alpha, h)) \cdot \mathbbm{1}_{\{t+h=\text{E} \}} }{\sum_{y_{i,t}\in \tilde{\Omega}} \sum^{H}_{h=1} |y_{i}| }, \quad \alpha = 50,90 .
\label{eqn:metric_hve}
\end{equation}
Here, E represents a peak event, and the metric is specifically measured against forecasts generated $h$ weeks apart from the target date, i.e., the event date. In this paper, we consider two major yearly promotional events, denoted as Event A and Event B. 
We omit the normalized mean absolute error (NMAE), as it is similar to the $P_{50}$ quantile loss. 
Other evaluation metrics, such as normalized mean square error (NMSE), are omitted, as they receive less interest and are not employed when evaluating both previous E-Commerce Retail studies~\cite{eisenach2020mqtransformer}.


\subsubsection{Reproducibility.} 
All models are trained using eight NVIDIA Tesla V100 32GB GPUs. 
Detailed information on implementation, hyperparameter tuning, and training efficiency is provided in Appendix~\ref{sec:appb}.
All evaluations are averaged over three independent runs or random seeds, and the standard deviation is reported in Appendix~\ref{sec:appc} for methods with close performance. 
For privacy considerations, results on the E-Commerce Retail Datasets have been normalized relative to RNN results (e.g., RNN results are set to 1.0, and other results are scaled accordingly).

\subsection{Demand Forecasting Results} 

Here, we present the evaluations of forecasting results for \method and baseline models on E-Commerce Retail Datasets. Since the primary objective is to evaluate peak forecasting performance, we pick two major sales events shared by two datasets, and we compare all models' performance in terms of target date accuracy (equation (\ref{eqn:metric_hve})) by setting the target date as an event date. In addition, we also compare overall accuracy across all models. 


\input{tables/04_overall_events.tex}

\subsubsection{Peak forecast accuracy.}
\label{sec:eva_peak}
We evaluate the forecasting performance over promotional events.
We fix the forecast generation date and evaluate model performance in terms of the target date accuracy, and we focus on two events, Event A and B, occurring at the \(3^{\text{rd}}\) and \(23^{\text{rd}}\) horizons, respectively. 
This evaluation follows a standard time-series forecasting test paradigm~\cite{lim2021temporal}. 
The evaluation results are summarized in Table~\ref{tbl:fixdate}. The results show clear advantages of \method in improving peak demand forecasting accuracy, particularly for shorter horizons. \method achieves up to 10\% lower \(P_{50}\) error and 20\% lower \(P_{90}\) error for Event A and ~5\% lower \(P_{50}\) and \(P_{90}\) errors on Event B. 
Furthermore, the performance improvements in E-Commerce Retail-Region 1 are more significant than in Region 2. 
This can be attributed to the fact that E-Commerce Retail-Region 1 exhibits sharper demand peaks during these events, where \method is expected to be more effective.


\subsubsection{Overall forecast accuracy.}
\label{sec:eva_avg}

In addition to the performance gains on forecasting demand peaks, we also evaluate the performance of \method to show competitive performance compared to state-of-the-art baselines on overall forecasting accuracy across all horizons.
We follow standard time-series forecasting setups~\cite{zhou2021informer, lim2021temporal}, where predictions are made at a fixed date, same as Table~\ref{tbl:fixdate}, and extend up to 26 weeks horizons. 
The results that are aggregated across all forecasting horizons are summarized in Table~\ref{tbl:retail_all}. 
The proposed \method demonstrates competitive performance compared to state-of-the-art demand forecasting models, achieving leading results on \(P_{50}\) errors and comparable performance on \(P_{90}\) errors for both the E-Commerce Retail-Region 1 and Region 2 datasets.


\subsection{Ablation Study}



\begin{figure}[ht]
    \centering
    \includegraphics[width=0.9\linewidth]{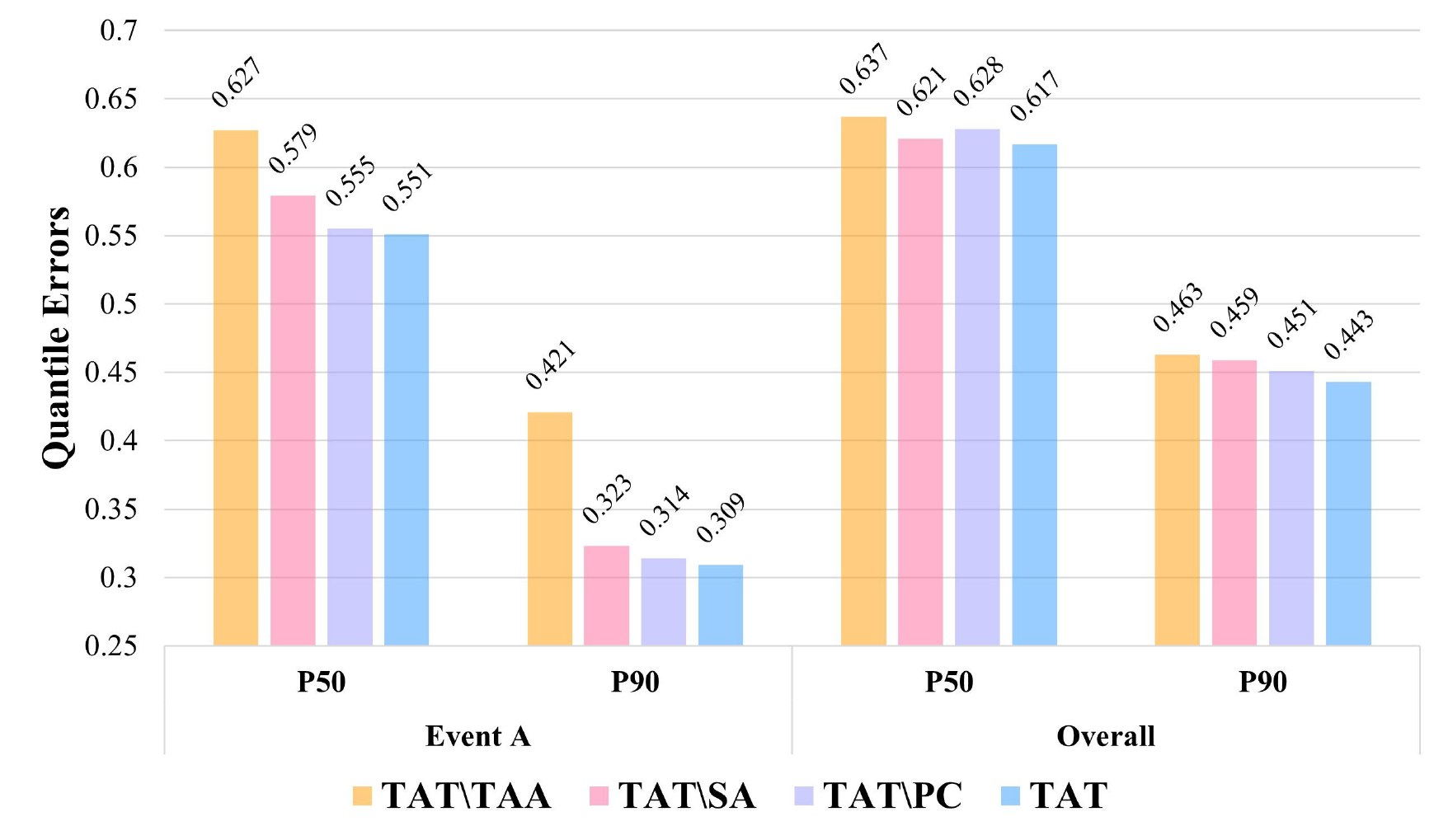}
    \Description{}
    \caption{Ablation study on \method and its three variants. The results demonstrate the effectiveness of the proposed \att, which achieves incremental improvements of 12\% in P50 and 26\% in P90 during Event A. Furthermore, removing any module from \method results in a performance decline, underscoring the importance of other modules to the effectiveness of \method.}
    \label{fig:ablation}
\end{figure}

We conduct an ablation study to evaluate the contribution of each module in \method. 
Specifically, we compare the performance of \method with three ablated variants: \methodns$\setminus$TAA, which excludes the \att in both encoder and decoder; \methodns$\setminus$SA, which excludes the self-attention in both encoder and decoder; and \methodns$\setminus$PC, which excludes the posterior calibration. 
The evaluations are performed on the E-Commerce Retail-Region 1 dataset, analyzing both the overall performance and the performance during Event A. 
The results of the ablation study are presented in Figure~\ref{fig:ablation}.


The results of the ablation study demonstrate the effectiveness of each module in \method, as evidenced by the performance degradation observed in all ablated variants, where one module is removed. 
Notably, excluding the \att module (\methodns$\setminus$TAA) results in the most significant performance drop for Event A demand peak predictions, with approximately 15\% higher $P_{50}$ error and 40\% higher $P_{90}$ error, in spite of the overall performance remains relatively close to that of \method. 
This finding highlights the critical role of the proposed \att module in aligning and modeling the correlations between future features and demand time series, thereby enhancing the accuracy of peak demand predictions.

%% file: tables/01_promotional_events.tex
\begin{table*}[ht]
\caption{Evaluations with fixed forecast generation date for two main peak events on two proprietary demand datasets. The best results are bold, and the second-best results are underlined. All values indicate relative performances versus the RNN~result.
}
\bgroup
\begin{tabular}{ccccccccccccccc}
\hline
\multicolumn{12}{c}{\texttt{\textbf{E-Commerce Retail-Region 1}}} \\
Event & Horizon & Metric & RNN & N-BEATS & PatchTST & iTrans. & MQCNN & TFT & MQT & TSMixer & \method \\ \hline
\multirow{2}{*}{Event A} & \multirow{2}{*}{3} & P50 & 1.000 & 0.685 & 0.667 & 0.674 & 0.700 & 0.623 & \underline{0.602} & 0.632 & \textbf{0.551} \\
& & P90 & 1.000 & 0.551 & 0.505 & 0.520 & 0.651 & 0.454 & \underline{0.380} & 0.445 & \textbf{0.309} \\ \hline
\multirow{2}{*}{Event B} & \multirow{2}{*}{23} & P50 & 1.000 & 0.768 & 0.746 & 0.835 & 0.742 & 0.759 & 0.740 & \underline{0.736} & \textbf{0.717} \\
& & P90 & 1.000 & 0.582 & 0.570 & 0.645 & 0.552 & 0.650 & 0.556 & \underline{0.541} & \textbf{0.529} \\ \hline
\multicolumn{12}{c}{\texttt{\textbf{E-Commerce Retail-Region 2}}} \\
Event & Horizon & Metric & RNN & N-BEATS & PatchTST & iTrans. & MQCNN & TFT & MQT & TSMixer & \method \\ \hline
\multirow{2}{*}{Event A} & \multirow{2}{*}{3} & P50 & 1.000 & 1.100 & 0.892 & 0.923 & 0.918 & 0.865 & \underline{0.835} & 0.935 & \textbf{0.830} \\
& & P90 & 1.000 & 0.766 & 0.635 & 0.690 & 0.770 & 0.585 & \underline{0.554} & 0.665 & \textbf{0.503} \\ \hline
\multirow{2}{*}{Event B} & \multirow{2}{*}{23} & P50 & 1.000 & 0.951 & 0.542 & 0.544 & 0.534 & 0.884 & \underline{0.528} & 0.800 & \textbf{0.513} \\
& & P90 & 1.000 & 0.894 & 0.761 & 0.766 & \underline{0.641} & 0.912 & \textbf{0.639} & 0.776 & 0.659 \\ \hline
\end{tabular}
\label{tbl:fixdate}
\egroup
\end{table*}

%% file: tables/04_overall_events.tex
\begin{table*}[ht]
\caption{Overall accuracy metrics aggregated across all horizons on two proprietary demand datasets. 
The best results are bold, and the second-best results are underlined. All values are relative performances to the RNN~result.}
\bgroup
\begin{tabular}{ccccccccccccc}
\hline
\multicolumn{10}{c}{\texttt{\textbf{E-Commerce Retail-Region 1}}} \\
Metric & RNN & N-BEATS & PatchTST & iTrans. & MQCNN & TFT & MQT & TSMixer & \method \\ \hline
P50 & 1.000 & 0.669 & 0.705 & 0.643 & 0.624 & 0.653 & \textbf{0.611} & 0.653 & \underline{0.617} \\
P90 & 1.000 & 0.520 & 0.542 & 0.520 & 0.479 & 0.538 & \underline{0.452} & 0.479 & \textbf{0.443} \\ \hline
\multicolumn{10}{c}{\texttt{\textbf{E-Commerce Retail-Region 2}}} \\
Metric & RNN & N-BEATS & PatchTST & iTrans. & MQCNN & TFT & MQT & TSMixer & \method \\ \hline
P50 & 1.000 & 0.999 & 0.645 & 0.659 & 0.630 & 0.896 & \underline{0.624} & 0.880 & \textbf{0.617} \\
P90 & 1.000 & 0.869 & 0.745 & 0.766 & \underline{0.675} & 0.840 & \textbf{0.669} & 0.769 & 0.681 \\ \hline
\end{tabular}
\label{tbl:retail_all}
\egroup
\end{table*}

%% file: content/06_conclusion.tex
\section{Conclusion}

In this work, we have proposed a novel transformer-based framework, \method, aimed at improving demand forecasting accuracy during demand peaks by effectively leveraging known context features. 
\method uses a novel attention mechanism, \att, which aligns and models the correlations between future features and demand time series during peak periods, resulting in more accurate predictions. 
In addition, the newly proposed posterior calibration in \method further enhances performance by directly calibrating decoder outputs and known context features. 
Empirical evaluations demonstrate the effectiveness of \method in improving forecasting accuracy during demand peaks, while maintaining competitive performance compared to state-of-the-art models in terms of overall  forecasting accuracy.





%% file: content/02_relate.tex
\section{Related Work}
\label{sec:relate}

\par\noindent\textbf{Time-Series Forecasting.} 
Classical statistical time-series forecasting models, such as ARIMA~\cite{hyndman2024forecasting}, often face limitations in capturing complicated patterns and dependencies due to inherent model constraints~\citep{nadaraya1964estimating, williams1995gaussian, smola2004tutorial}. 
Recent work that applied machine learning to time-series forecasting using RNNs, LSTNet, N-BEATS has achieved notable improvements in accuracy~\cite{chung2014empirical, lai2018modeling, oreshkin2019n, olivares2022nbeatsx}. 
State-of-the-art models build upon the successes of transformer-based architectures~\cite{vaswani2017attention}, such as Informer, Autoformer, Fedformer, PatchTST, iTransformer~\cite{zhou2021informer, wu2021autoformer, zhou2022fedformer, Yuqietal-2023-PatchTST, liu2023itransformer}, have shown significantly improved forecasting accuracy. In parallel, a growing body of work explores model-agnostic strategies to enhance out-of-distribution generalization in time-series forecasting, further boosting performance independently of the underlying forecasting model~\cite{zhao2025performative, liu2024time2, kim2021reversible, liu2023adaptive, fan2023dish}.

\par\noindent\textbf{Demand Forecasting.} 
General time-series forecasting has achieved significant success, but demand forecasting is a variant of time-series forecasting that requires a setup that differs from standard approaches. 
Specifically, it involves predicting targets by leveraging exogenous features, some (but not all) of which are known in advance. 
Early efforts focused on conventional recurrent models~\cite{wen2017multi, Olivares2021ProbabilisticHF}, while recent advancements explored transformer-based models and foundation models, achieving notable improvements in forecasting accuracy~\cite{lim2021temporal, eisenach2020mqtransformer, yang2022mqret, yang2023geann, kamarthi2023large, das2023decoder}. 
Additionally, the rise of multimodal approaches in time-series forecasting~\cite{jin2023time, liu2024lstprompt, liutime, liu2024picture} has inspired methods that integrate unstructured text data for demand forecasting~\cite{zhangllmforecaster}. 
However, these approaches~\cite{lim2021temporal, zhangllmforecaster} often face challenges in data and computational scalability, making it difficult to apply them effectively to real-world, large-scale demand forecasting datasets.

%% file: content/08_appb.tex
\section{Empirical Evaluation Detail}
\label{sec:appb}

In this section, we provide additional details on our empirical evaluation.

\subsection{Dataset Statistics}

We here present the dataset statistics for the dataset used in this work, including two industrial datasets from the e-commerce retailer. The statistics are shown in Table~\ref{tbl:stat}.

\input{tables/03_amazon_datasets.tex}




\subsection{Model Implementation}  
For the industrial e-commerce retail datasets, we set the forecasting horizon window size to $H=26$ (half year), aligning with industrial requirements and interests~\cite{wen2017multi, eisenach2020mqtransformer}. The lookback window size $L$ varies depending on the model. For MQCNN and MQT, we use the full historical data ($L=208$) as the lookback window. For other sequential models, we use a shorter lookback window of $L=104$. This difference is based on the characteristics of the dataset and models. Specifically, certain demand time series exhibit low values at earlier stages that rise sharply later. Using an excessively long lookback window for sequential models may cause instability in fitting these cases. Consequently, a shorter lookback window is preferred for these models. In contrast, MQCNN and MQT mitigate this limitation by leveraging importance sampling, which allows them to effectively use full historical data ($L=208$).  


For baseline implementations, we follow the publicly available implementations provided by established libraries. Specifically, we use the N-BEATS implementation from the repository available at \url{https://github.com/philipperemy/n-beats}, the TFT implementation from\url{https://github.com/google-research/google-research/tree/master/tft}, and the transformer-based model implementations from Time-Series Library at \url{https://github.com/thuml/Time-Series-Library}. These repositories offer reliable implementations, ensuring consistency and reproducibility in baseline comparisons.

\subsection{Hyperparameter Tuning} 


We tune the model hyperparameters following different strategies on each dataset, ensuring alignment with the specific characteristics of the data. For the E-Commerce Retail datasets, we follow the common practices established in prior studies~\cite{eisenach2020mqtransformer}. Specifically, we employ batch sizes of 64 and epochs of 10 on MQCNN and MQT with importance sampling, and batch size 512 and epochs of 50 or 100 epochs for other sequential models without importance sampling. Additional hyperparameters such as hidden size and number of layers are fine-tuned using a small subset of the training data. 


\begin{table}[!t]
\centering
\caption{\method's hyperparameters for E-Comm. datasets.}
\label{tbl:hyper}
\begin{tabular}{ccc}
\toprule
\textbf{Hyperparameter}     & \textbf{Region 1} & \textbf{Region 2} \\
\midrule
Hidden                      & 60                        & 60                        \\
Dropout ($x_s$)             & 0.5                       & 0.5                       \\
Dropout (Other)             & 0.1                       & 0.1                       \\
Heads                       & 1                         & 1                         \\
Optimizer                   & AdamW                     & AdamW                     \\
Batchsize                   & 512                       & 512                       \\
Learning Rate               & 1e-3                      & 1e-3                      \\
Epochs                      & 100                       & 50                        \\
Earlystop                   & \ding{55}                 & \ding{55}                         \\
\bottomrule
\end{tabular}
\end{table}

The hyperparameters used in \method are detailed in Table~\ref{tbl:hyper}. All experiments are conducted using three random seeds or three independent runs, and the reported results are averaged across these runs. 

\subsection{Training Efficiency} 
The training efficiency on the E-Commerce Retail Datasets of selected models used in our experiments is reported in Table~\ref{tbl:time}. Models with undesirable performance, such as RNN, are excluded from this analysis due to their limited relevance and lower practical interest. As shown in the table, \method demonstrates reasonable training efficiency compared to other baseline models. While it trains slightly slower than general time-series forecasting models like TSMixer, it significantly outperforms practical demand forecasting models such as MQT (which also gives more accurate predictions in most cases) in terms of training speed.


\begin{table}[!t]
\caption{Training efficiency between \method and baselines on E-Commerce Retail Datasets.}
\begin{tabular}{cccc}
\hline
Method & Batchsize & Epoch & Efficiency \\ \hline 
PatchTST & 512 & 100 & 0.5x \\ 
TSMixer & 512 & 100 & 0.4x \\ 
TFT & 64 & 50 & 11.4x \\ 
MQCNN & 64 & 100 & 2.3x \\ 
MQT & 64 & 100 & 9.0x \\ 
\method & 512 & \begin{tabular}[c]{@{}c@{}}50 (E) \\ 100 (A)\end{tabular} & \begin{tabular}[c]{@{}c@{}}1.0x \\ ($\sim$8 sec/epoch)\end{tabular} \\ \hline
\end{tabular}
\label{tbl:time}
\end{table}

%% file: tables/03_amazon_datasets.tex
\begin{table}[h]
\centering
\caption{Statistics of the datasets used in the evaluations. "Unique" indicates that each sample is from distinct entities with no overlap.}
\label{tbl:stat}
\begin{tabular}{ccc}
\toprule
\textbf{Dataset}       & \textbf{Region 1} & \textbf{Region 2} \\
\midrule
Train Samples          & 270k                      & 280k                      \\
Test Samples           & 280k                      & 340k                      \\
Entities               & Unique                    & Unique                    \\
Lookback Window        & 208                       & 208                       \\
Forecast Horizon       & 26                        & 26                        \\
\bottomrule
\end{tabular}
\end{table}

%% file: content/09_appc.tex


\begin{table}[!t]
\caption{Standard deviations for \method and baseline models with close performance on E-Commerce Retail Datasets (Performance on demand peaks with fixed prediction date, extended Table~\ref{tbl:fixdate}).}
\begin{tabular}{cccccc}
\hline
\multicolumn{6}{c}{\texttt{\textbf{E-Commerce Retail-Region 1}}} \\
Event & Metric & MQCNN & MQT & TSMixer & \method \\ \hline
\multirow{4}{*}{A} & \multirow{2}{*}{P50} & $0.700$ & $\underline{0.602}$ & $0.632$ & $\textbf{0.551}$ \\
& & (\small{$\pm$0.0066}) & (\small{$\pm$0.0130}) & (\small{$\pm$0.0121}) & (\small{$\pm$0.0132}) \\ \cline{2-6}
& \multirow{2}{*}{P90} & $0.651$ & $\underline{0.380}$ & $0.445$ & $\textbf{0.309}$ \\ 
& & (\small{$\pm$0.0068}) & (\small{$\pm$0.0184}) & (\small{$\pm$0.0129}) & (\small{$\pm$0.0198}) \\ \hline
\multirow{4}{*}{B} & \multirow{2}{*}{P50} & $0.742$ & $0.740$ & $\underline{0.736}$ & $\textbf{0.717}$ \\
& & (\small{$\pm$0.0053}) & (\small{$\pm$0.0128}) & (\small{$\pm$0.0072}) & (\small{$\pm$0.0115}) \\ \cline{2-6}
& \multirow{2}{*}{P90} & $0.552$ & $0.556$ & $\underline{0.541}$ & $\textbf{0.529}$ \\ 
& & (\small{$\pm$0.0053}) & (\small{$\pm$0.0180}) & (\small{$\pm$0.0130}) & (\small{$\pm$0.0098}) \\ \hline
\multicolumn{6}{c}{\texttt{\textbf{E-Commerce Retail-Region 2}}} \\
Event & Metric & MQCNN & MQT & TSMixer & \method \\ \hline
\multirow{4}{*}{A} & \multirow{2}{*}{P50} & $0.918$ & $\underline{0.835}$ & $0.935$ & $\textbf{0.830}$ \\
& & (\small{$\pm$0.0008}) & (\small{$\pm$0.0147}) & (\small{$\pm$0.0165}) & (\small{$\pm$0.0020}) \\ \cline{2-6}
& \multirow{2}{*}{P90} & $0.770$ & $\underline{0.554}$ & $0.665$ & $\textbf{0.503}$ \\ 
& & (\small{$\pm$0.0049}) & (\small{$\pm$0.0180}) & (\small{$\pm$0.0110}) & (\small{$\pm$0.0059}) \\ \hline
\multirow{4}{*}{B} & \multirow{2}{*}{P50} & $0.534$ & $\underline{0.528}$ & $0.800$ & $\textbf{0.513}$ \\
& & (\small{$\pm$0.0100}) & (\small{$\pm$0.0086}) & (\small{$\pm$0.0220}) & (\small{$\pm$0.0068}) \\ \cline{2-6}
& \multirow{2}{*}{P90} & $\underline{0.641}$ & $\textbf{0.639}$ & $0.776$ & $0.659$ \\ 
& & (\small{$\pm$0.0155}) & (\small{$\pm$0.0166}) & (\small{$\pm$0.0197}) & (\small{$\pm$0.0111}) \\ \hline
\end{tabular}
\label{tbl:fixdate_std}
\end{table}

\begin{table}[!t]
\caption{Standard deviations for \method and baseline models with close performance on E-Commerce Retail Datasets (Averaged performance on all horizons, extended Table~\ref{tbl:retail_all}).}
\begin{tabular}{ccccc}
\hline
\multicolumn{5}{c}{\texttt{\textbf{E-Commerce Retail-Region 1}}} \\
Metric & MQCNN & MQT & TSMixer & \method \\ \hline
\multirow{2}{*}{P50} & $0.624$ & $\textbf{0.611}$ & $0.653$ & $\underline{0.617}$ \\
& (\small{$\pm$0.0014}) & (\small{$\pm$0.0036}) & (\small{$\pm$0.0140}) & (\small{$\pm$0.0040}) \\ \hline
\multirow{2}{*}{P90} & $0.479$ & $\underline{0.452}$ & $0.479$ & $\textbf{0.443}$ \\ 
& (\small{$\pm$0.0021}) & (\small{$\pm$0.0062}) & (\small{$\pm$0.0066}) & (\small{$\pm$0.0053}) \\ \hline
\multicolumn{5}{c}{\texttt{\textbf{E-Commerce Retail-Region 2}}} \\
Metric & MQCNN & MQT & TSMixer & \method \\ \hline
\multirow{2}{*}{P50} & $0.630$ & $\underline{0.624}$ & $0.880$ & $\textbf{0.617}$ \\ 
& (\small{$\pm$0.0042}) & (\small{$\pm$0.0056}) & (\small{$\pm$0.0190}) & (\small{$\pm$0.0064}) \\ \hline
\multirow{2}{*}{P90} & \underline{0.675} & $\textbf{0.669}$ & $0.769$ & $0.681$ \\
& (\small{$\pm$0.0041}) & (\small{$\pm$0.0139}) & (\small{$\pm$0.0048}) &(\small{$\pm$0.0100}) \\ \hline
\end{tabular}
\label{tbl:retail_std}
\end{table}



\section{Additional Results}
\label{sec:appc}

To mitigate potential bias and noise in the results, we report the average performance of \method and baseline models across three independent runs with different random seeds in the main results. 
We further report the standard deviations of MQCNN, MQT, TSMixer, and \method of evaluations on the E-Commerce Retail Datasets (Table~\ref{tbl:fixdate} 
and Table~\ref{tbl:retail_all}), due to their close performance. 
(See the tables below.)
We choose these baselines as they have achieved the best or second-best results at least one time in the corresponding evaluations. Other baselines are excluded from these tables as their performances are uniformly lagging behind. 